\documentclass{article}

    \PassOptionsToPackage{numbers, compress}{natbib}

     \usepackage[preprint]{neurips_2024}

\usepackage[utf8]{inputenc} 
\usepackage[T1]{fontenc}    
\usepackage{hyperref}       
\usepackage{url}            
\usepackage{booktabs}       
\usepackage{amsfonts}       
\usepackage{nicefrac}       
\usepackage{microtype}      
\usepackage{xcolor}         

\usepackage{times}
\usepackage{epsfig}
\usepackage{graphicx}
\usepackage{amsmath}
\usepackage{amssymb}
\usepackage{enumitem}
\usepackage{bbm}
\usepackage{wrapfig}

\usepackage{multicol}
\usepackage{colortbl}
\usepackage{multirow}
\usepackage{pifont}
\usepackage{makecell}
\usepackage[ruled,linesnumbered]{algorithm2e}

\usepackage{amsmath}

\title{SyncNoise: Geometrically Consistent Noise Prediction for Text-based 3D Scene Editing}

\author{%
  {\small Ruihuang Li$^{1,2}$
   \; 
   Liyi Chen$^1$ 
   \;
   Zhengqiang Zhang$^{1,2}$
   \;
   Varun Jampani$^3$ 
   \;
   Vishal M. Patel$^4$
   \;
   Lei Zhang$^{1,2}$\footnotemark[1]\thanks{Corresponding author. This work is supported by the PolyU-OPPO Joint Innovation Lab.}}\\
  {\footnotesize $^1$Hong Kong Polytechnic University\; $^2$OPPO Research Institute\; $^3$Stability AI\; $^4$Johns Hopkins University} \\
 {\small \texttt{\{csrhli, cslzhang\}@comp.polyu.edu.hk \; vpatel36@jhu.edu } }\\
  \href{https://lslrh.github.io/syncnoise.github.io/}{Project website} 
}

\begin{document}

\maketitle
\vspace{-6mm}
\begin{figure}[!h]
	\centering 
	\includegraphics[scale=0.19]{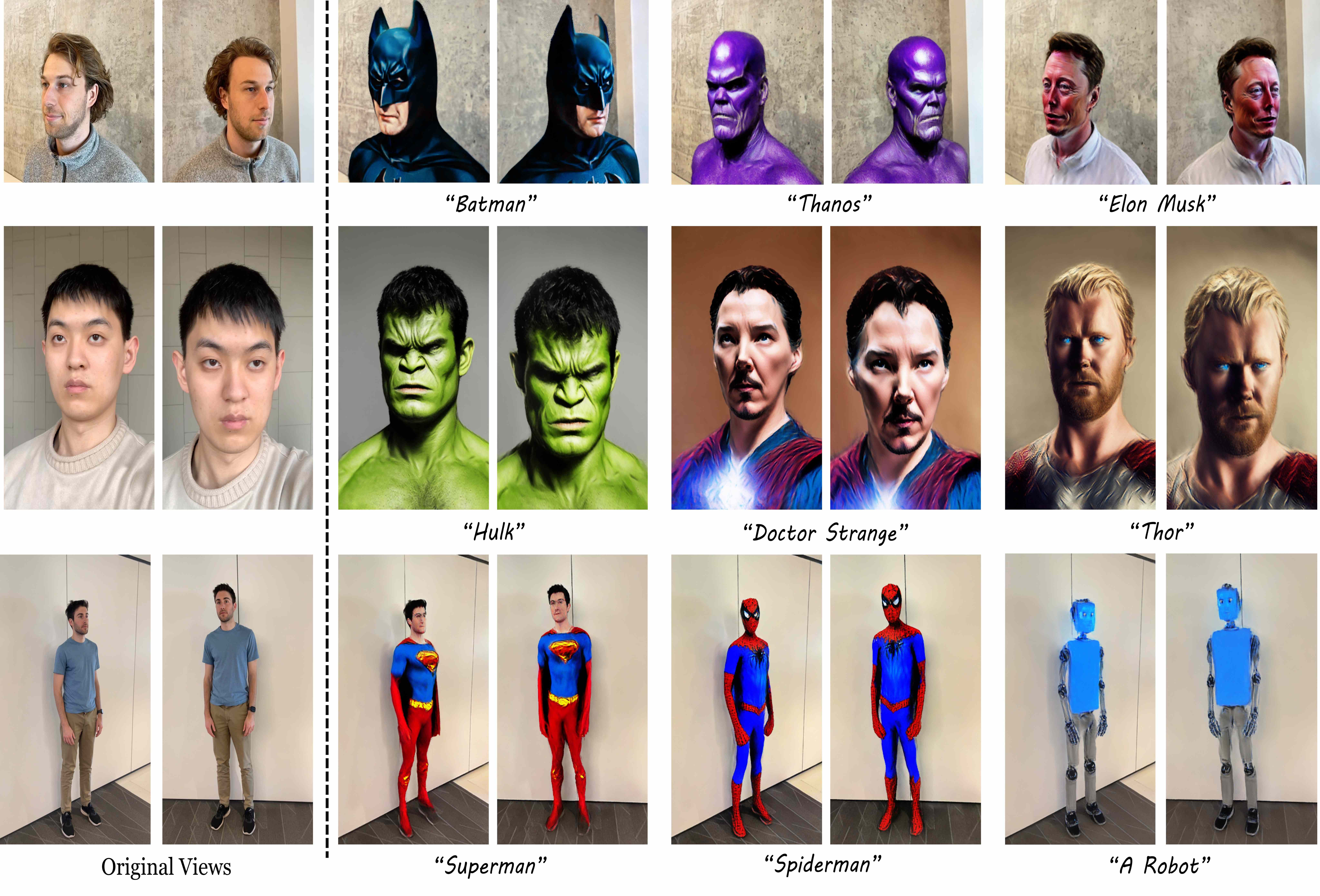}\\
	\vspace{-2mm}
	\caption{Edited results by \textbf{SyncNoise}, which achieves high-quality and controllable editing that closely adheres to the instructions with minimal changes to irrelevant regions. SyncNoise attains geometrically consistent editing without compromising fine-grained textures.}
	\label{results} 
\end{figure}

\begin{abstract}
Text-based 2D diffusion models have demonstrated impressive capabilities in image generation and editing. Meanwhile, the 2D diffusion models also exhibit substantial potentials for 3D editing tasks. However, how to achieve consistent edits across multiple viewpoints remains a challenge. While the iterative dataset update method is capable of achieving global consistency, it suffers from slow convergence and over-smoothed textures. We propose SyncNoise, a novel geometry-guided multi-view consistent noise editing approach for high-fidelity 3D scene editing. SyncNoise synchronously edits multiple views with 2D diffusion models while enforcing multi-view noise predictions to be geometrically consistent, which ensures global consistency in both semantic structure and low-frequency appearance. To further enhance local consistency in high-frequency details, we set a group of anchor views and propagate them to their neighboring frames through cross-view reprojection. To improve the reliability of multi-view correspondences, we introduce depth supervision during training to enhance the reconstruction of precise geometries. Our method achieves high-quality 3D editing results respecting the textual instructions, especially in scenes with complex textures, by enhancing geometric consistency at the noise and pixel levels.   
\end{abstract}

\section{Introduction}
Text-based 3D scene editing is an emerging field that focuses on creating and manipulating 3D scenes using natural language instructions. Given an original 3D representation, one can achieve a wide variety of edits using abundant and flexible textual instructions, such as modifying the geometry, appearance, lighting, textures, and other attributes of the scene to achieve desired effects or fulfill design objectives. Despite the advancements in 3D generative diffusion models~\cite{hong2023lrm,wang2024crm}, it still requires a significant amount of paired 3D scene data to adapt these models for 3D editing tasks. Given the limited availability of such data, an alternative approach is to distill the prior knowledge from 2D diffusion models to improve 3D representations. 

Diffusion-based image editing approaches, including text-driven image synthesis and editing~\cite{rombach2022high, hertz2022prompt}, stroke-based editing~\cite{meng2021sdedit}, exemplar-based methods~\cite{yang2023paint}, and point-based editing~\cite{shi2023dragdiffusion} have achieved considerable success and facilitated artistic creation. Despite the increasing maturity and accessibility of 3D reconstruction techniques~\cite{mildenhall2021nerf,kerbl20233d}, applying 2D editing strategies to 3D scenes has not been extensively studied in the literature.
One straightforward solution is to utilize a 2D diffusion model to edit each view separately, and then use the edited multi-view images to update the 3D representations to obtain the desired shapes and textures. However, due to the inherent randomness of diffusion process and the lack of 3D priors, it is challenging for a 2D model to generate multi-view consistent editing results in terms of geometry, lighting, and textures simultaneously. 

To alleviate this issue, Instruct-Nerf2Nerf~\cite{haque2023instruct} (IN2N) presents an iterative dataset update framework to alternatively edit one randomly selected view with InstructPix2Pix~\cite{brooks2023instructpix2pix} and optimize the 3D scenes based on the edited image. Although IN2N can achieve globally consistent editing, it suffers from \textit{longer optimization duration} to obtain a satisfactory edited scene. Besides, it \textit{eliminates fine-grained details} that are not consistent across views, leading to over-smoothed results. 

To improve the editing efficiency, Efficient-Nerf2Nerf~\cite{song2023efficient} (EN2N) incorporates multi-view consistency regularization into the diffusion process and achieves consistent outputs in a single pass. However, this approach suffers from blurry results for two reasons. Firstly, optical flow is employed for cross-view matching, which may lead to \textit{imprecise correspondences}, especially when the view changes significantly. Secondly, EN2N imposes a consistency constraint on the latent codes of diffusion models, which tends to collapse the rich and nuanced latent representation into a more averaged form, consequently leading to a loss of high-frequency details and subtle variations that are critical for realistic editing.

In order to avoid the blurred editing results and generate finer-grained textures, in this paper we propose \textbf{SyncNoise}, a geometry-aware multi-view synchronized noise prediction method for 3D scene editing. Firstly, we leverage geometric information of 3D scenes to achieve precise and dense multi-view matching, which paves the way for applying multi-view consistency constraints at the noise and pixel levels. Since implicit 3D representations, such as Neural Radiance Field (NeRF) models, often suffer from unreliable geometry fitting, we introduce additional depth supervision produced by running Structure-from-Motion (SfM)~\cite{schoenberger2016sfm,schoenberger2016mvs} to improve the geometric reconstruction, avoiding aligning non-matched regions of different views. 

Secondly, motivated by the observation that intermediate features of the noise predictor (U-Net) not only involve semantic information but also exhibit the \textit{structure-to-appearance controllability}~\cite{zhang2024tale,liu2024drag,voynov2023p+}, we \textit{enforce multi-view consistency on the U-Net features for predicting noise maps}, rather than on the latent map. This not only effectively mitigates the smoothed results by performing average operations on the latent map, but also \textbf{achieves multi-view consistent edits in semantic structure and low-frequency appearance}. Since solely manipulating the noise predictions cannot ensure consistent high-frequency details across adjacent views, we further \textit{employ a cross-view projection strategy to propagate the anchor views to others} for improving the pixel-level consistency. Fig.~\ref{results} shows some editing results on different 3D scenes. We can observe that by leveraging the geometric information to synchronously predict multi-view noise maps, and propagating well-edited view to its neighboring views, our proposed SyncNoise can achieve consistent and efficient 3D edits respecting the textual instructions and retain more details in edited scenes.

\section{Related Work}\vspace{-3mm}
\textbf{Image Generation and Editing.}
Recently, diffusion models~\cite{ho2020denoising} have demonstrated excellent semantic understanding capability for image generation. Conditioned on a given textual prompt,
DALL-E-2~\cite{ramesh2022hierarchical} and Stable Diffusion~\cite{rombach2022high} achieve impressive generation performance using classifier-free guidance.
Most image editing methods inherit the prior knowledge of pre-trained generation models  to modify the appearance and shapes of reference images while preserving their original structure. Prompt2Prompt~\cite{hertz2022prompt} (P2P) aligns the source attention maps of source and edited images with the given text prompts to achieve localized editing. InstructPix2Pix~\cite{brooks2023instructpix2pix} extends P2P to support instruction-based efficient editing. 
The following work Plug-and-Play~\cite{tumanyan2023plug} injects the reference feature and self-attention layer to control the editing process.  
Delta denoising score (DDS)~\cite{hertz2023delta} extends score distillation sampling (SDS) to avoid significant background changes.
Personalized generation shares similarity with image editing.
Textural inversion~\cite{gal2022image}  expands the language-vision dictionary to inject the subject content into a word embedding to achieve subject-driven generation.
Similarly, DreamBooth~\cite{ruiz2023dreambooth} inverses the subject by finetuning the Stable Diffusion to achieve better fidelity.
The following studies~\cite{mokady2023null,li2024source} improve the editing quality by preserving the context from the inversion process.
These 2D editing studies provide a good starting point for 3D editing. 

\begin{figure}[!t]
	\centering 
	\vspace{-2mm}
	\includegraphics[scale=0.34]{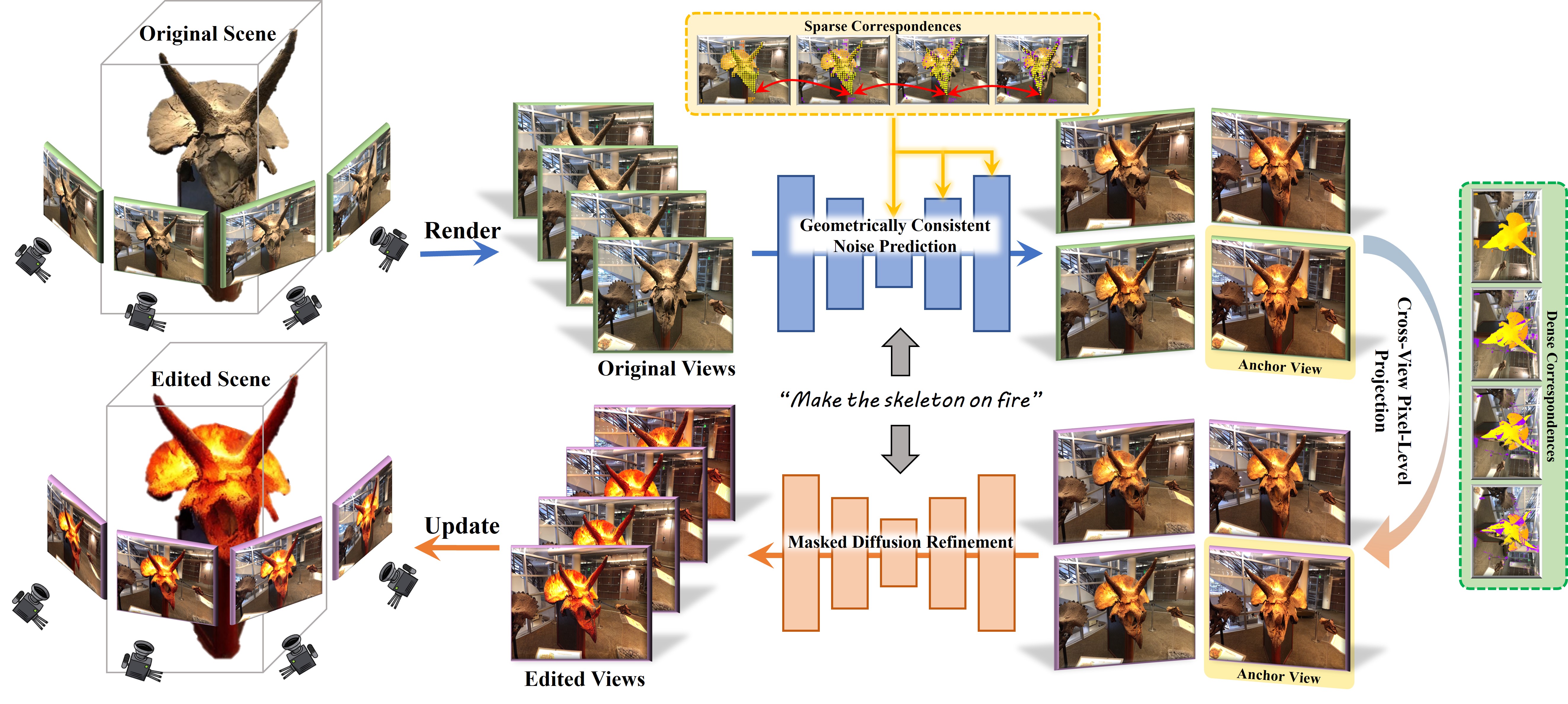}\\
	\vspace{-3mm}
	\caption{Overview of our proposed SyncNoise for text-based 3D scene editing. We edit rendered multi-view images while enforcing geometrical consistency at the noise and pixel levels. First, we construct reliable correspondences based on precise 3D geometries. Then, we enforce multi-view noise consistency by aligning U-Net decoder features across views. We also use cross-view projection to maintain pixel-level consistency by propagating the anchor view to neighboring views. To minimize reprojection artifacts, we refine these views with a 2D diffusion model. Finally, we update the 3D scene based on the edited multi-view images.}
	\label{pipeline} 
	\vspace{-4mm}
\end{figure}

\textbf{3D Scene Editing.} Many studies have explored editing neural fields in different manners.  SKED~\cite{mikaeili2023sked} and Liu et.al.~\cite{liu2021editing} employ 2D sketches to control editing.
Some works attempt to manipulate NeRF utilizing point clouds~\cite{chen2023neuraleditor}, feature volumes~\cite{lazova2023control}, attributes~\cite{kania2022conerf}, or mesh~\cite{jambon2023nerfshop, yuan2022nerf, yang2022neumesh}. However, these methods demand complex human interactions, limiting their applications. 
Driven by the development of LLMs and multi-modality models~\cite{radford2021learning, rombach2022high}, instruction-based editing has attracted much attention due to its user-friendly nature.
NeRF-Art~\cite{wang2023nerf} and ClipNeRF~\cite{wang2022clip} edit global NeRF by maximizing the CLIP similarity between rendered 2D views and text prompt. 
FocalDreamer~\cite{li2024focaldreamer} and Instruct-3Dto3D~\cite{kamata2023instruct} optimize the 3D models using SDS loss~\cite{poole2022dreamfusion} from pre-trained Stable Diffusion and Instruct-Pix2Pix, respectively.
Similarly, Shap-Editor~\cite{chen2023shap} learns a feed-forward network to directly output the edited NeRF latent. 
To enable fine-grained localized editing, Distilled Feature Fields~\cite{kobayashi2022decomposing} and Neural Feature Fusion Fields~\cite{tschernezki2022neural} introduce pre-trained 2D models of LSeg~\cite{li2022language} and DINO~\cite{caron2021emerging} for localization. 
More recently, Instruct-Nerf2Nerf~\cite{haque2023instruct} iteratively updates 3D model and edits rendered images. 
DreamEditor~\cite{zhuang2023dreameditor} leverages DreamBooth~\cite{ruiz2023dreambooth} for subject-driven editing under the given text prompt without sacrificing the fidelity to the original object.
GenN2N~\cite{liu2024genn2n} distills the priors from off-the-shelf 2D models in latent space to achieve 3D editing.
GaussianEditor~\cite{chen2023gaussianeditor} maintains a dynamic mask for localized editing based on 3D Gaussians. 
DGE~\cite{chen2024dge} reformulates 3D editing as a video editing task~\cite{wu2023tune, qi2023fatezero} by injecting temporal attention to capture 3D consistency.
However, these methods omit the depth cues, resulting in inconsistent editing results. In contrast, our method alters to multi-view consistent noise prediction based on depth-aware correspondences.


\section{Method}
In this work, we focus on text-based 3D scene editing by resorting to 2D diffusion models. Given an original 3D representation (NeRF or Gaussian Splatting), multi-view images and their corresponding camera poses, we aim to produce an edited scene under the guidance of natural-language instructions.

As shown in Fig~\ref{pipeline}, we leverage instruction-based 2D diffusion models to edit multi-view images, followed by optimizing the original 3D representations using the edited views as supervision. Ensuring multi-view consistent editing is crucial, as any inconsistencies in textures between views can lead to undesirable smoothing effects. To this end, we first leverage 3D geometry to establish precise multi-view correspondences in Sec.~\ref{sec2.1}. Secondly, in Sec.~\ref{sec2.2}, we impose multi-view consistency constraint on the noise predictions throughout the denoising (editing) process, for enhancing semantic and appearance coherence across the views. Furthermore, to preserve more high-frequency details, we employ cross-view projection to propagate the editing effects from anchor views to their neighboring views in Sec.~\ref{sec2.3}, so as to achieve multi-view consistent edits at the pixel level. 

\begin{figure}[!t]
	\centering 
	\includegraphics[scale=0.37]{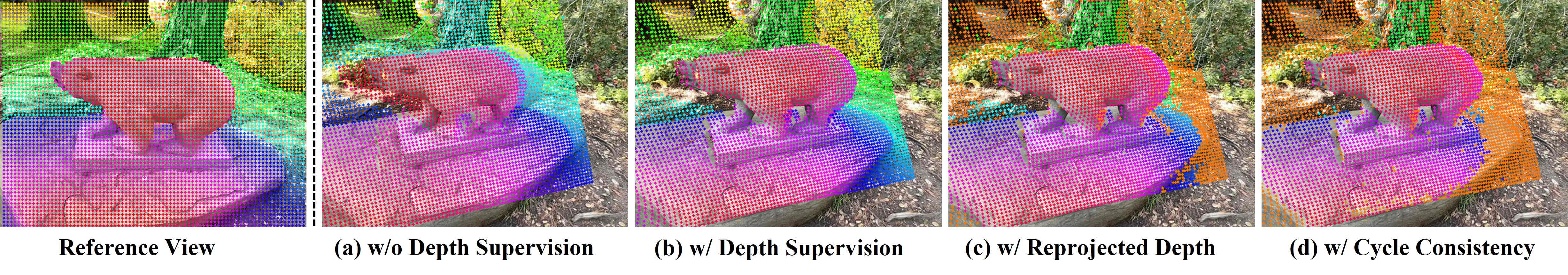}\\
	\vspace{-2mm}
	\caption{The estimated depth on reference view $D_{ref}$ and the re-projected depth from reference view to novel view $D_{ref\to k}$. By imposing the depth supervision and two constraints, we can obtain reliable geometric correspondences across views. \textcolor{orange}{\textbf{Orange}} denotes noisy points to be filtered.}
	\label{depth} 
	\vspace{-3mm}
\end{figure}   
\subsection{Reliable Geometry-guided Correspondence}
\label{sec2.1}
To establish reliable correspondences among multiple views, we incorporate depth supervision to enhance the reconstructed geometry. Furthermore, we leverage the re-projected depth and cycle consistency constraints to filter out unreliable matching points, ensuring the matching accuracy.

{\bf Depth Supervision.} The implicit 3D representation, such as NeRF, exhibits limited capability in fitting geometry, particularly in scenarios with sparse views. Consequently, the predicted depth by NeRF tends to be unreliable. As shown in Fig.~\ref{depth}(a), there are significant offsets when reprojecting points from reference view to others. To address this limitation, we follow~\cite{deng2022depth} to introduce depth supervision into the training process of NeRF. Specifically, we derive the depth supervision from 3D keypoints obtained by running Structure-form-Motion (SfM) solver~\cite{schoenberger2016sfm}, and add a depth loss to enforce the estimated depth to match the depth of keypoints. As shown in Fig.~\ref{depth}(b), by adding the depth supervision, we are able to estimate more precise depth, which in turn enables us to establish dense and accurate correspondences among different views.  

{\bf Reprojected Depth Constraint.} While explicit depth supervision can enhance the quality of 3D geometry, there are still deviations in the matching points due to the noise of 3D key points. To filter out noisy correspondences, we compare the reprojected depth from the reference view $I_{ref}$ to the $k$-th novel view $I_{k}$, denoted by $D_{ref\to k}$, with the estimated depth on the $k$-th view, denoted by $D_{k}$, and retain the matching points that  satisfy the following condition:
\begin{align}
	|D_{ref\to k}-D_{k}| < \tau_{d},
\end{align}
where $\tau_{d}$ denotes the depth threshold used to eliminate noisy matching points. As can be observed in Fig.~\ref{depth}(c), most of background points, occluded points in the novel view, and points located at the edges of objects have been filtered out. As the span between views increases, the number of reliable matching points gradually decreases. 

{\bf Cycle Consistency Constraint.} In addition to the reprojected depth constraint, reliable matching points should also adhere to cycle consistency constraint. The pixel distance between the point back-projected from $I_{k}$ to $I_{ref}$, denoted by $P_{ref\to k\to ref}$, and its original starting point, denoted by $P_{ref}$, should satisfy the following condition:
\begin{align}
	|P_{ref\to k\to ref}-P_{ref}| < \tau_{p},
\end{align}
where $|\cdot|$ calculates the pixel distance between two points, and $\tau_{p}$ is the threshold used to filter out noisy points that can not be back-projected to their original locations. \textit{Please refer to Appendix~\ref{app:cycle} for more details about cycle consistency constraint.}
 
 \begin{figure}[!t]
 	\centering 
 	\includegraphics[scale=0.2]{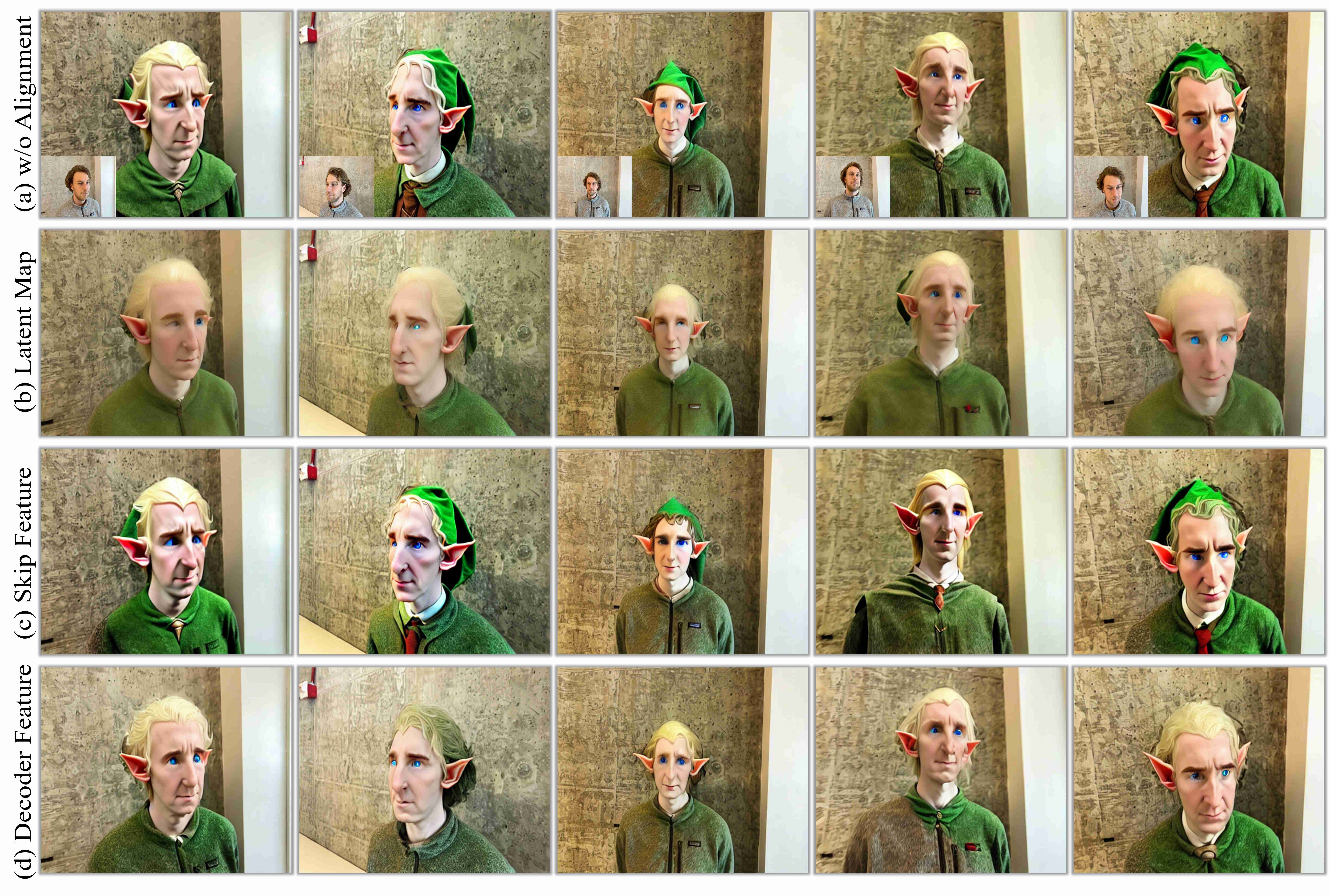}\\
 	\vspace{-4mm}
 	\caption{Multi-view editing results obtained (a) without alignment, (b) by aligning latent features of different views, by enforcing consistencies on (c) skip features and (d) decoder features of U-Net. By enforcing the decoder features of noise predictor to be consistent, we can obtain multi-view consistent edits without introducing blurs. The text prompt is ``\textit{make the man look like Tolkien Elf}''.}
 	\label{noise} 
 	\vspace{-4mm}
 \end{figure}  
\subsection{Geometrically Consistent Noise Prediction}
\label{sec2.2}
Building upon the precise geometric correspondences we constructed in Sec.~\ref{sec2.1}, in this section we aim to enforce the editing results from multiple views to be consistent throughout the whole denoising process from $T$ to $0$ steps. A simple and effective approach to achieve this goal is by averaging the corresponding latent features across multiple views~\cite{song2023efficient}. However, this method has two major limitations. On one hand, directly manipulating the latent maps can lead to smoothed results in generated images, as shown in Fig.~\ref{noise}(b). On the other hand, assigning equal weights to different views is not reasonable due to varying qualities of matching points across views, making the model biased to views with poor correspondences. 

Prior studies~\cite{zhang2024tale,voynov2023p+,liu2024drag} have demonstrated that the intermediate features of noise predictor (U-Net) not only capture semantic information but also influence the final appearance of image. This motivates us to enhance multi-view consistency on the U-Net features rather than latent maps. In Fig.~\ref{noise}(d), it can be observed that by enforcing consistency on the intermediate decoder features of U-Net, we can achieve multi-view consistent editing results without introducing blurred artifacts. When the constraint is applied to skip features, the impact is relatively minor, as shown in Fig~\ref{noise}(c). \textit{Please refer to Appendix.~\ref{app:alignment} for the architecture of U-Net and the effects of aligning different layers of U-Net. }

{\bf Initial Noise Alignment.} We first align the initial noise from multiple views. Specifically, given the random noise from $K$ different views, denoted by $\{Z^1_T,\cdots,Z^K_T\}$, and the correspondences among them, we define the noise of each reference view as the weighted sum of noises from $K$ views:
\begin{align}
	Z_{T}^{ref,i} = \sum_{k=1}^{K} w_k^i \cdot \mathbbm{1}_{match}^{k,i} \cdot Z_{T}^{k,i},
\end{align}
where $Z_{T}^{ref,i}$ denotes the noise vector of the $i$-th point in the reference view. $\mathbbm{1}^{k,i}_{match}$ is an indicator function that equals $1$ if the $k$-th view contains a matching point for the $i$-th point in the reference view. $w_k^i$ represents the weight assigned to the $k$-th view, which is inversely proportional to the reprojection error, denoted by $\delta_{k}^i=|D_{ref\to k}-D_k|$. The weight is defined as follows:
\begin{align}
	w_k^i = e^{-\mu\cdot \delta_{k}^i},
\end{align}
where $\mu$ controls the relative gaps between weights on different views. Each weight is normalized by the sum of weights for all matching points.

{\bf Masked Multi-View Consistent Noise Editing.} 
In addition to initial random noise, we also align the noise predictions across all diffusion steps $t\in [0,\cdots,T]$. Given the $l$-th layer features of noise predictor from $K$ different views, denoted by $\{F_l^1,\cdots,F_l^{K}\}, l\in\{1,\cdots,11\}$, we aggregate multi-view noise features corresponding to the $i$-th point into the $I_{ref}$ through the following formula:
\begin{align}
	F_{l}^{ref,i} = \sum_{k=1}^{K} w_k^i \cdot \mathbbm{1}_{match}^{k,i} \cdot F_{l}^{k,i}.
\end{align} 

Furthermore, to achieve more precise foreground editing without modifying irrelevant regions, we introduce masks to restrict the matching and editing regions. We retain only the correspondences within the mask, and filter out redundant associations from unrelated regions. In addition, during each denoising step, we apply a mask to limit the region of text guidance and modify the noise estimate equation as follows (\textit{please refer to Appendix~\ref{app:cfg} for more details about this equation.}):
\begin{equation}
	\begin{split}
	\hat{\epsilon}_{\theta}(z_t,c_I,c_T) &= \epsilon_{\theta}(z_t, \varnothing, \varnothing)+g_I\cdot (\epsilon_{\theta}(z_t,c_I,\varnothing)-\epsilon_{\theta}(z_t,\varnothing,\varnothing))\\&+g_T\cdot(\epsilon_{\theta}(z_t,c_I,c_T)-\epsilon_{\theta}(z_t,c_I,\varnothing))\cdot M_{soft},
	\end{split}
\label{eq6}
\end{equation} 

 \begin{figure}[!t]
	\centering 
	\includegraphics[scale=0.2]{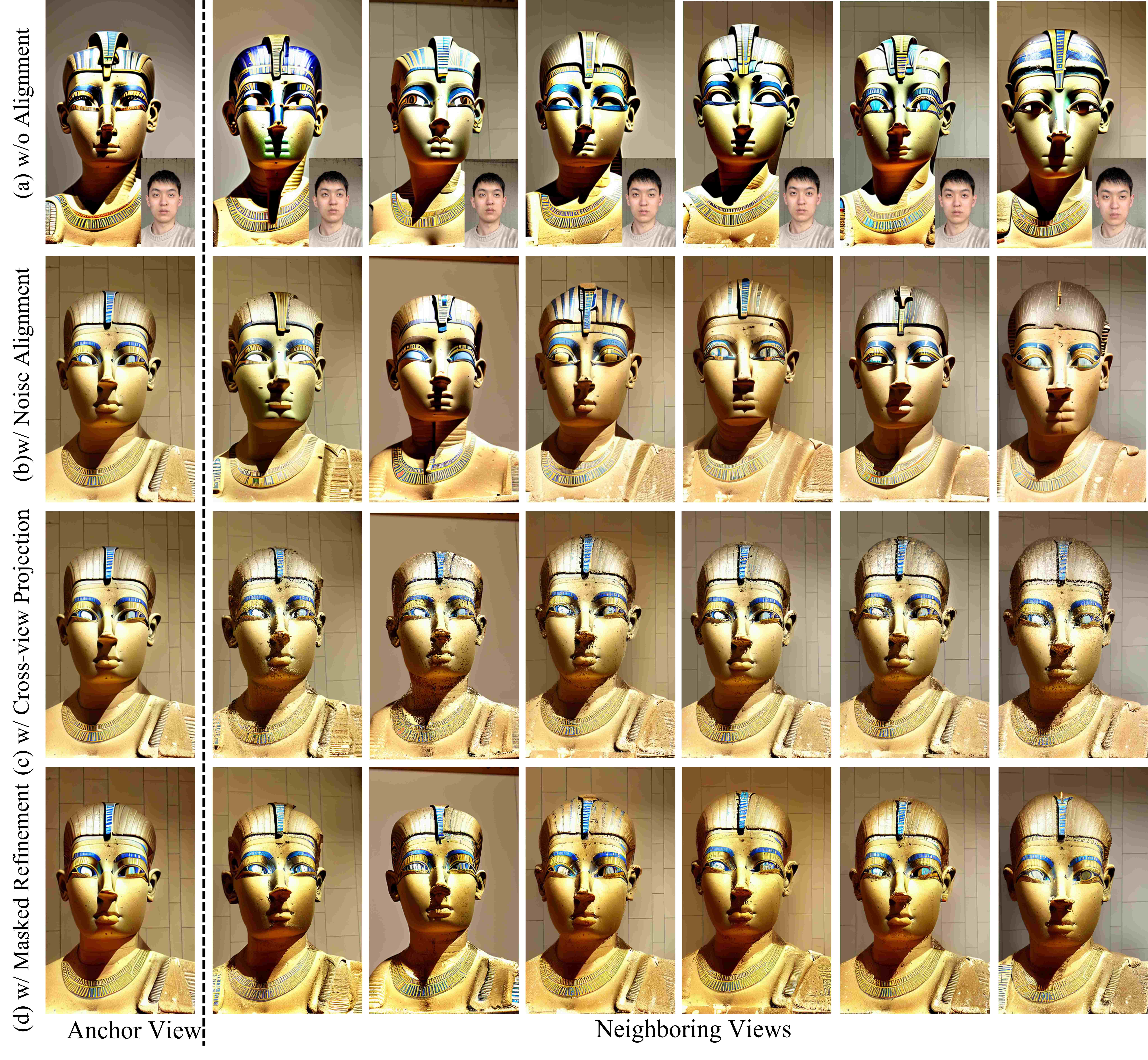}\\
	\vspace{-4mm}
	\caption{Multi-view editing results. Noise alignment is responsible for producing consistent edits in semantic structure and low-frequency appearance, while cross-view pixel reprojection ensures the consistency in high-frequency details. The text prompt is ``\textit{Turn him into an Egyptian sculpture}''.}
	\label{pixel} 
	\vspace{-4mm}
\end{figure} 

where $c_{I}$, $c_T$ and $\varnothing$ denote the image, text, and no conditions, respectively. $\epsilon_{\theta}$ is the denoising U-Net. $g_I$ and $g_T$ are two classifier-free guidance scales for balancing the quality and diversity of samples generated by the diffusion model. It is worth noting that we employ a \textbf{soft mask}, denoted by $M_{soft}$, instead of a binary mask for masked noise estimation. This is because \textit{trivially reducing the background weights to zero would also decrease the editing fidelity on foreground}. Specifically, the weights of foreground regions are set to $1$, while the weights of backgrounds gradually decay from $0.5$ to $0$ as they move away from the center of foreground.

\subsection{Cross-View Pixel-Level Projection}
\label{sec2.3}
We have aligned the initial noise and noise predictions of U-Net from multiple views, which can achieve globally consistent edits in a more efficient manner than the iterative refinement strategy~\cite{haque2023instruct}. However, as shown in Fig.~\ref{pixel}(b), noise-level alignment can only ensure consistency in semantic structure and low-frequency textures, but cannot guarantee consistency in high-frequency details. Even a small misalignment in these details can ultimately result in smoothed textures in 3D edits.

{\bf Cross-View Projection.} To preserve more fine-grained textures in the edited 3D scenes, we need to perform pixel-level alignment. Specifically, we propose to propagate partial anchor views to others based on the established dense correspondences in Sec.~\ref{sec2.1}. First, we utilize metrics such as CLIP directional similarity score to select well-edited views as the anchors, denoted by $I^e_{anchor}$. Then we reproject each $I^e_{anchor}$ to its neighboring views $I^e_{k}$ and use the reprojected pixels to replace the corresponding pixels in $I^e_{k}$:
\begin{align}
	I^e_{k}[M^{val}_{k}] = I^e_{anchor}[M^{val}_{anchor}],	
\end{align}
where $M^{val}_{k}$ and $M^{val}_{anchor}$ indicate valid correspondences satisfying the depth and cycle consistency constraints in $I^e_{k}$ and $I^e_{anchor}$, respectively. 

{\bf Masked Diffusion Refinement.}
As shown in Fig.~\ref{pixel}(c), cross-view reprojection further improves the consistency in fine-grained details across adjacent views. However, it may cause artifacts in novel views. To address this issue, we further perform masked refinement by feeding the edited view $I_{k}^e$ into the 2D editing model as follows:
\begin{equation}
	\begin{split}
		\hat{\epsilon}_{\theta}(z_t,c_{I^e_k},c_T) &= \epsilon_{\theta}(z_t, \varnothing, \varnothing)+g_I\cdot (\epsilon_{\theta}(z_t,c_{I^e_k},\varnothing)-\epsilon_{\theta}(z_t,\varnothing,\varnothing))\\&+g_T\cdot(\epsilon_{\theta}(z_t,c_{I^e_k},c_T)-\epsilon_{\theta}(z_t,c_{I^e_k},\varnothing))\cdot (M_{k}-M_{k}^{val}).
	\end{split}
\end{equation}
Unlike Eq.~\ref{eq6}, in the above equation, the edited image rather than the original image is used as image condition. Additionally, the text guidance is only applied on the regions not replaced by anchor views, which are represented by $(M_{k}^{fore}-M_{k}^{val})$, where $M_{k}^{fore}$ denotes the mask of foreground object. As can be seen in Fig.~\ref{pixel}(d), through masked refinement, the prior knowledge from $I_{anchor}^e$ is incorporated into the unprojected regions of $I_{k}^e$.

\begin{figure}[!t]
	\centering 
	\includegraphics[scale=0.47]{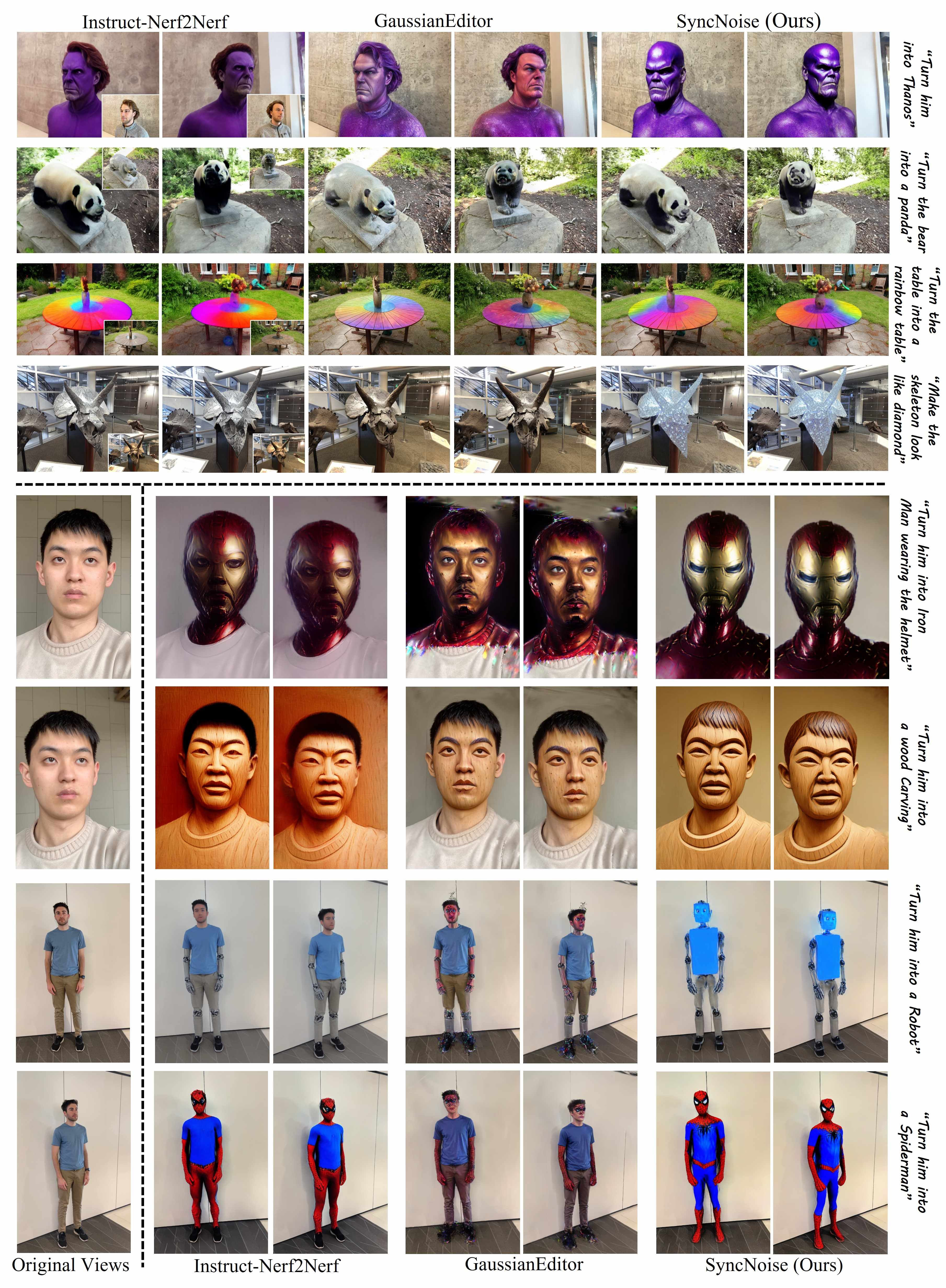}\\
	\vspace{-3mm}
	\caption{Qualitative comparisons. Our SyncNoise offers more consistent (\textit{e.g.} ``rainbow table''), finer-grained (\textit{e.g.} ``wood carving'', ``Spiderman''), and instruction-following 3D editing (\textit{e.g.} ``Iron Man wearing the helmet'', ``robot'', ``Thanos'') with minimal changes to irrelevant regions. }
	\label{compare} 
	\vspace{-0.3cm}
\end{figure} 
\subsection{3D Representation Optimization} 
The proposed multi-view synchronized noise prediction achieves consistent edits in both structure and appearance, while the cross-view pixel-level projection further enhances the consistency among neighboring views. Based on the edited results of all views, we first train the 3D model for 1000-2000 iterations, depending on the complexity of the scenes, to inject the 2D edits into 3D representation. Subsequently, we employ an iterative refinement~\cite{haque2023instruct} approach to further enhance the 3D representation. Note that our approach differs from IN2N~\cite{haque2023instruct} in a key aspect. In IN2N, during the early optimization steps, the multi-view image edits exhibit significant inconsistencies, leading to over-smoothed 3D edits. In contrast, our method first generates multi-view consistent 2D edits to ensure general consistency in 3D geometry and appearance, and then employs an iterative refinement process to adjust finer details.

\section{Experiments}
\textbf{Implementation Details.} 
During the editing process, we first edit $80$ multi-view images while enforcing consistency on the layer-5 and layer-8 of U-Net features (see Fig.~\ref{app:unet} in the Appendix). Subsequently, for the anchor view selection, we pick the view with the highest CLIP direction score in every 10 adjacent views as the anchor view, and reproject them onto neighboring views with about 80\% overlap. \textit{Please refer to Appendix.~\ref{app:setup} for more implementation details.}


\textbf{Evaluation.} We use three metrics to measure the alignment to textual instruction, \textit{i.e.}, CLIP similarity score, CLIP text-image directional similarity~\cite{brooks2023instructpix2pix}, and CLIP temporal directional similarity~\cite{haque2023instruct}. In addition, we employ two no-reference image quality assessment metrics, \textit{i.e.}, CLIP-IQA~\cite{wang2023exploring} and MUSIQ~\cite{ke2021musiq}, for evaluation. \textit{Please refer to Appendix.~\ref{app:setup} for the detailed information about metrics.}

\subsection{Qualitative Results}
In Fig.~\ref{results}, we demonstrate some edits with different text prompts. As can be observed from the edits with prompts ``Batman'' and ``Robot'', our method still exhibits multi-view consistency even when the geometry and shape of original scenes undergo obvious changes. Additionally, we can see finer details in the hair of ``Hulk'', arms of ``Spiderman'', and clothing of ``Thor''. This is because we enforce consistency on adjacent frames through pixel-level projection. \textit{Please see the supplemental video to better evaluate the quality and consistency of our results.}


We compare our proposed SyncNoise with two representative instruction-based methods, Instruct-Nerf2Nerf~\cite{haque2023instruct} and GaussianEditor~\cite{fang2023gaussianeditor} in Fig.~\ref{compare}. We reproduce the results of compared methods with their official codes and default parameters. Our SyncNoise achieves realistic and consistent edits that are faithful to the input textual instruction. In the example of ``Turn the table into a rainbow table'', our edits exhibit better multi-view \textbf{consistency} compared to the other two methods. IN2N exhibits color blending issues due to inconsistent edits in each iteration. For the instruction ``turn him into a wood carving'', our SyncNoise successfully edits even the hair and produces \textbf{fine-grained} textures. In addition, our results strictly \textbf{adhere to editing prompt} ``turn him into Iron Man wearing the helmet'', and generate highly realistic helmet. However, GuassianEditor hardly changes the appearance of human face, as it restricts the updates of old Gaussian points, hindering their editing fidelity to texts. Our method achieves superior edits by enforcing global structural and appearance consistency at the noise level, as well as improving local texture consistency at the pixel level. \textit{Please refer to Appendix.~\ref{app:qualitative} for more qualitative results.}

\begin{table}[!t]
	\caption{Quantitative evaluation. SyncNoise achieves high fidelity to the instruction prompt (CLIP score) without sacrificing the visual quality (MUSIQ), outperforming other methods across all metrics.}
	\vspace{-2mm}
	\scalebox{0.8}{
		\setlength{\tabcolsep}{3.7pt}
		\begin{tabular}{c|c|cc|ccc|cc|c}
			\toprule
			Method                     & {\small 3D Model}              & {\small Noise} & {\small Pixel} & {\scriptsize \begin{tabular}[c]{@{}c@{}}CLIP \\ Score$\uparrow$\end{tabular}} & {\scriptsize \begin{tabular}[c]{@{}c@{}}CLIP Text-Image \\ Direction Similarity$\uparrow$\end{tabular}} & {\scriptsize \begin{tabular}[c]{@{}c@{}}CLIP Temporal\\ Direction Similarity$\uparrow$\end{tabular}} & {\small CLIP-IQA$\uparrow$} & {\small MUSIQ$\uparrow$} & {\small \begin{tabular}[c]{@{}c@{}}Avg. \\ Time\end{tabular}} \\ \hline \hline
			IN2N~\cite{haque2023instruct}                       & NeRF                  &                                                                     &                                                                    &   29.18\%                                                         &   16.49\%    &  90.12\%    &  0.489  & 64.729    & 57min     \\ 
			EN2N~\cite{song2023efficient}                       & NeRF                  &          &           &    28.46\%                                                        &   15.74\%   &90.47\%            &  0.496    &  63.853  &19min        \\ 
			GaussianEditor~\cite{fang2023gaussianeditor}             & GS                    &                                                                     &                                                                    &  26.55\%                                          &    17.04\%      &   88.42\%   &  0.511  &  64.425    & 11min       \\ \hline
			\multirow{3}{*}{SyncNoise} & \multirow{3}{*}{NeRF} &  \checkmark                                                                    &                                                                    &  30.64\%           &  17.93\%       &  91.42\%    &    \textbf{0.559}  & 65.901   &  \multirow{3}{*}{23min}   \\ 
			&                       &                                                                     &     \checkmark                                                               &  29.50\%    &   16.97\%      &   89.90\%   &   0.504  & 65.221       \\
			&                       &  \checkmark                                                                    &  \checkmark                                                                   &  \textbf{30.86\%}      & \textbf{18.31\%}    &  \textbf{92.04\%}    &   0.540 & \textbf{66.668}        \\ \bottomrule
		\end{tabular}
	}
	\label{tab:qualtitative}
	\vspace{-4mm}
\end{table}

\vspace{-1mm}
\subsection{Quantitative Comparison}\vspace{-1mm}
We provide the quantitative comparison results between SyncNoise and Instruct-Nerf2Nerf (IN2N)~\cite{haque2023instruct}, Efficient-Nerf2Nerf (EN2N) ~\cite{song2023efficient} and GaussianEditor~\cite{fang2023gaussianeditor} in Tab.~\ref{tab:qualtitative}. We evaluate all the compared methods on a total of four scenes (\textit{i.e.}, `bear', `face', `fangzhou' and `person') and 10 different text prompts.
One can see that our method achieves superior editing performance on not only editing fidelity but also visual quality. Our method achieves better instruction-following edits and better temporal consistency, compared to IN2N, while requiring only half the editing time. Besides, our method outperforms GaussianEditor by 1.27\% and 2.243 in terms of CLIP text-image direction similarity score and MUSIQ, respectively, which indicates that our edited images are much more faithful to the given instructions without sacrificing the visual quality. GaussianEditor limits the update of partial 3D Gaussians points of original scene so that it cannot adhere to the instructions very well. Additionally, by introducing pixel-level consistency, SyncNoise further enhances the fidelity to instruction and visual quality, achieving finer-grained editing details across different views. 

\begin{figure}[!h]
	\centering 
	\includegraphics[scale=0.17]{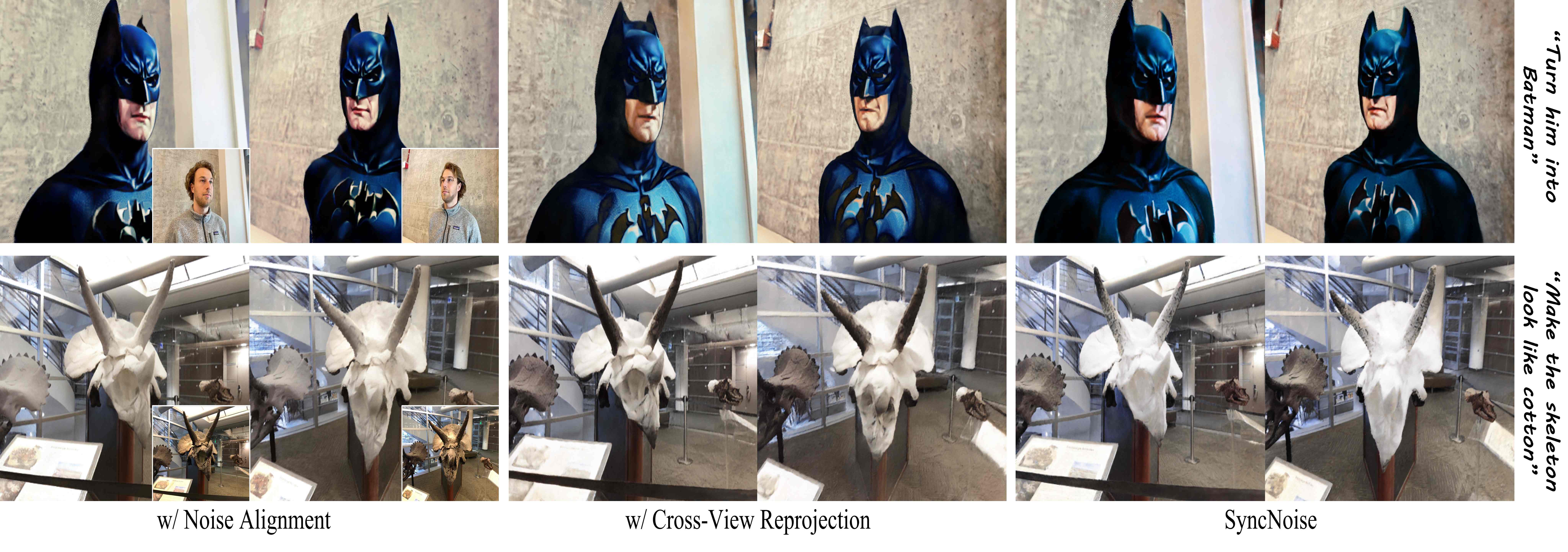}\\
	\vspace{-4mm}
	\caption{Ablation study on different components of our pipeline. Noise alignment achieves structural and general appearance consistency. Cross-view reprojection preserves more local details, but fails to maintain global coherence, \textit{e.g.}, the horns of skeleton are not edited.}
	\label{ablation} 
\end{figure} 

\vspace{-1mm}
\subsection{Ablation Study}\vspace{-1mm}
We investigate the roles of two key components in our pipeline, \textit{i.e.}, noise alignment and cross-view reprojection. As illustrated in Fig.~\ref{ablation}, with only the noise alignment, the edits are geometrically aligned but lose some details. With only the cross-view reprojection, the local consistency in adjacent frames is maintained, but the edits lack comprehensive coverage and completeness. For example, the horns of Batman are small, and the horns of the skeleton are not edited. By combining these two components together, we achieve better consistency in both global structure and local texture.
\vspace{-1mm}

\section{Conclusion}\vspace{-1mm}
In this work, we focused on achieving multi-view consistent edits in 3D scene editing. We proposed a novel approach called SyncNoise, which leveraged geometry-guided multi-view consistency to enhance the coherence of edited scenes. By synchronously editing multiple views using a 2D diffusion model and enforcing geometric consistency on the features of noise predictor, we avoided blurred outcomes. The pixel-level reprojection between neighboring views further helped generate more fine-grained details. Our experimental results demonstrated that SyncNoise outperformed existing methods in terms of achieving high-quality 3D editing while respecting textual instructions.

{\bf Limitations.} Our method heavily relies on 2D diffusion models, which limit the quality of 3D editing and the flexibility in prompts. Besides, SyncNoise excels in editing appearance but has limited capabilities in modifying 3D shape and geometry. This limitation is also inherent to IN2N. In addition, SyncNoise may struggle on complex scenes, especially those with intricate geometries. More investigations are required to address these challenges, which will be our future focuses.

{\small
	\bibliography{main}
	\bibliographystyle{plain}
}

\newpage
\appendix
\section{Appendix} 

\subsection{Cycle Consistency Constraint}  
\label{app:cycle} 
\begin{figure}[!h]
	\centering 
	\includegraphics[scale=0.7]{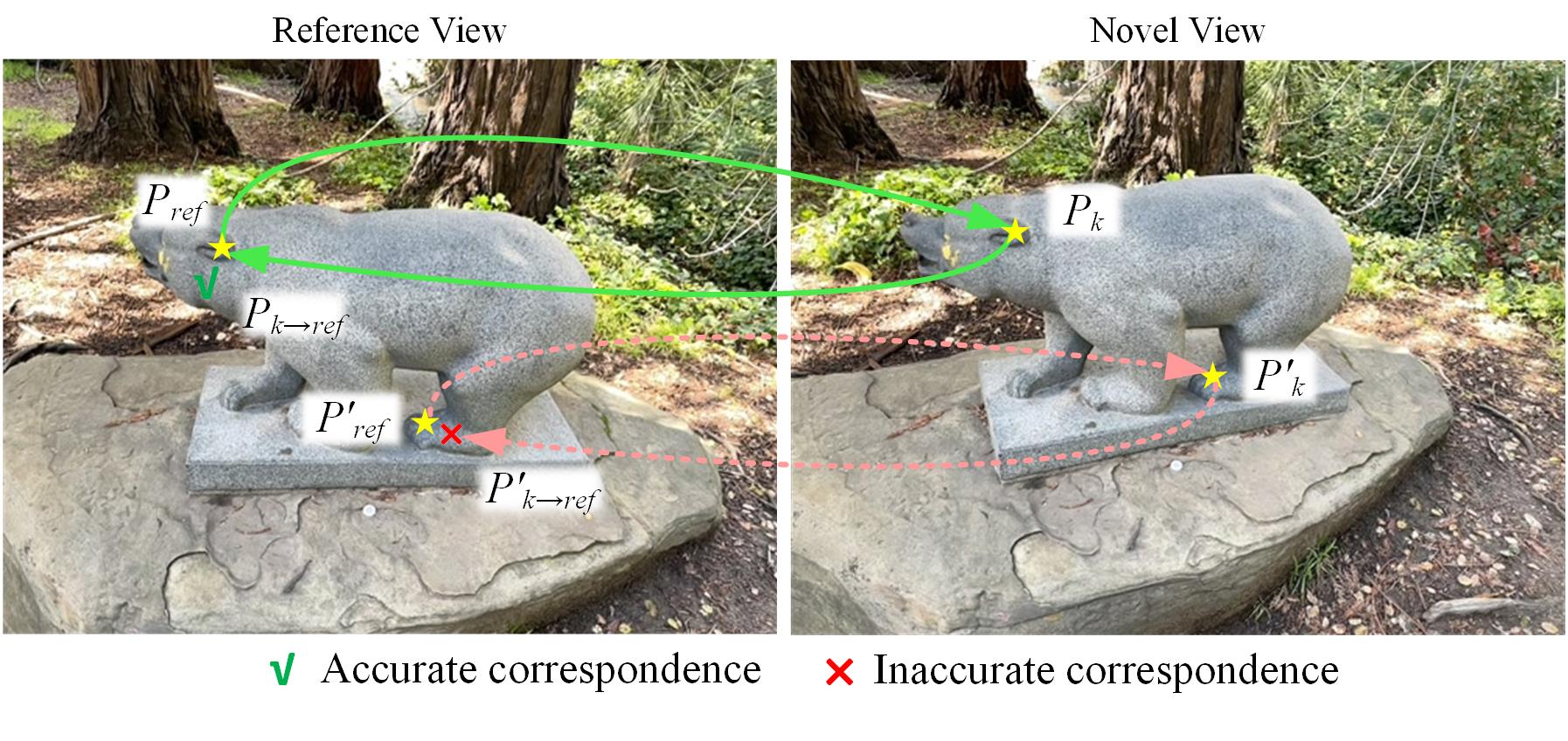}\\
	\vspace{-5mm}
	\caption{Cycle Consistency Constraint.}
	\label{cycle} 
\end{figure} 
We use the cycle consistency constraint to filter out partial noisy correspondences. The point $P_{k\to ref}$ which is back-projected from $I_{k}$ to $I_{ref}$, and its starting point $P_{ref}$, should be positioned in close pixel proximity. Symbol $\textcolor{green}{\checkmark}$ denotes the point satisfying the cycle consistency constraint, \textit{i.e.}, $|P_{ref}-P_{ref\to k\to ref}|<\tau_p$, which indicates that $P_{ref}$ and $P_{k}$ are accurately matched. Symbol $\textcolor{red}{\times}$ denotes the point that does not satisfy the constraint, indicating that the correspondence between $P'_k$ and $P'_{ref}$ is not precise.

\subsection{Classifier-Free Guidance}
\label{app:cfg}
Many prior works have demonstrated the effectiveness of classifier-free guidance (CFG) for integrating additional conditions during the inference process.
The objective of CFG is to shift the predicted scores towards locations that align more tightly with the conditions via extrapolation between unconditional scores and conditional scores.
In this paper, we follow InstructPix2Pix~\cite{brooks2023instructpix2pix} and employ a two-condition CFG strategy to ensure that the edited results are more faithful to the instructions and the original image.
The final extrapolated scores can be estimated as the follows:
\begin{equation}
	\begin{split}
		\hat{\epsilon}_{\theta}(z_t,c_I,c_T) &= \epsilon_{\theta}(z_t, \varnothing, \varnothing)+g_I\cdot (\epsilon_{\theta}(z_t,c_I,\varnothing)-\epsilon_{\theta}(z_t,\varnothing,\varnothing))\\&+g_T\cdot(\epsilon_{\theta}(z_t,c_I,c_T)-\epsilon_{\theta}(z_t,c_I,\varnothing)),
	\end{split}
\end{equation}
where $\varnothing$ is a fixed null value that represents the unconditional inputs, with $c_I$ and $c_T$ denoting the text instruction and input image, respectively. $g_I$ and $g_T$ are guidance weights designed to control the strength of the corresponding conditions.
During training, the conditions are randomly dropped to allow the denoising network $\epsilon_\theta$ to conduct denoising in a conditional or unconditional context for both or either conditional inputs.

\subsection{Experimental Setup}
\label{app:setup}
{\bf Implementation Details.} 
Our method is implemented based on the nerfstudio~\cite{tancik2023nerfstudio}. We use InstructPix2Pix~\cite{brooks2023instructpix2pix} as the image editor, which has been fine-tuned on image-to-image translation/editing dataset based on Stable Diffusion~\cite{rombach2022high}. As for the 3D dataset, we use four scenes (`\textit{bear}', `\textit{face}', `\textit{fangzhou}', and `\textit{person}') from IN2N~\cite{haque2023instruct}, the \textit{garden} scene from Mip-NeRF360~\cite{barron2022mip}, and other scenes from the LLFF~\cite{mildenhall2019local} datasets.  We employ the $L_1$ and LPIPS~\cite{zhang2018unreasonable} losses for optimizing 3D models. Experiments are conducted on one NVIDIA A100 GPU.

{\bf Evaluation.} We use three CLIP-related scores:
\textbf{CLIP similarity score} evaluates the alignment between text prompts and edited images.
\textbf{CLIP text-image directional similarity}~\cite{brooks2023instructpix2pix} evaluates the alignment between the change in text captions and the change in images.
\textbf{CLIP temporal directional similarity}~\cite{haque2023instruct} evaluates the temporal consistency in the CLIP space.

\subsection{Qualitative Comparison}
\label{app:qualitative}
\begin{figure}[!h]
	\centering 
	\includegraphics[scale=0.18]{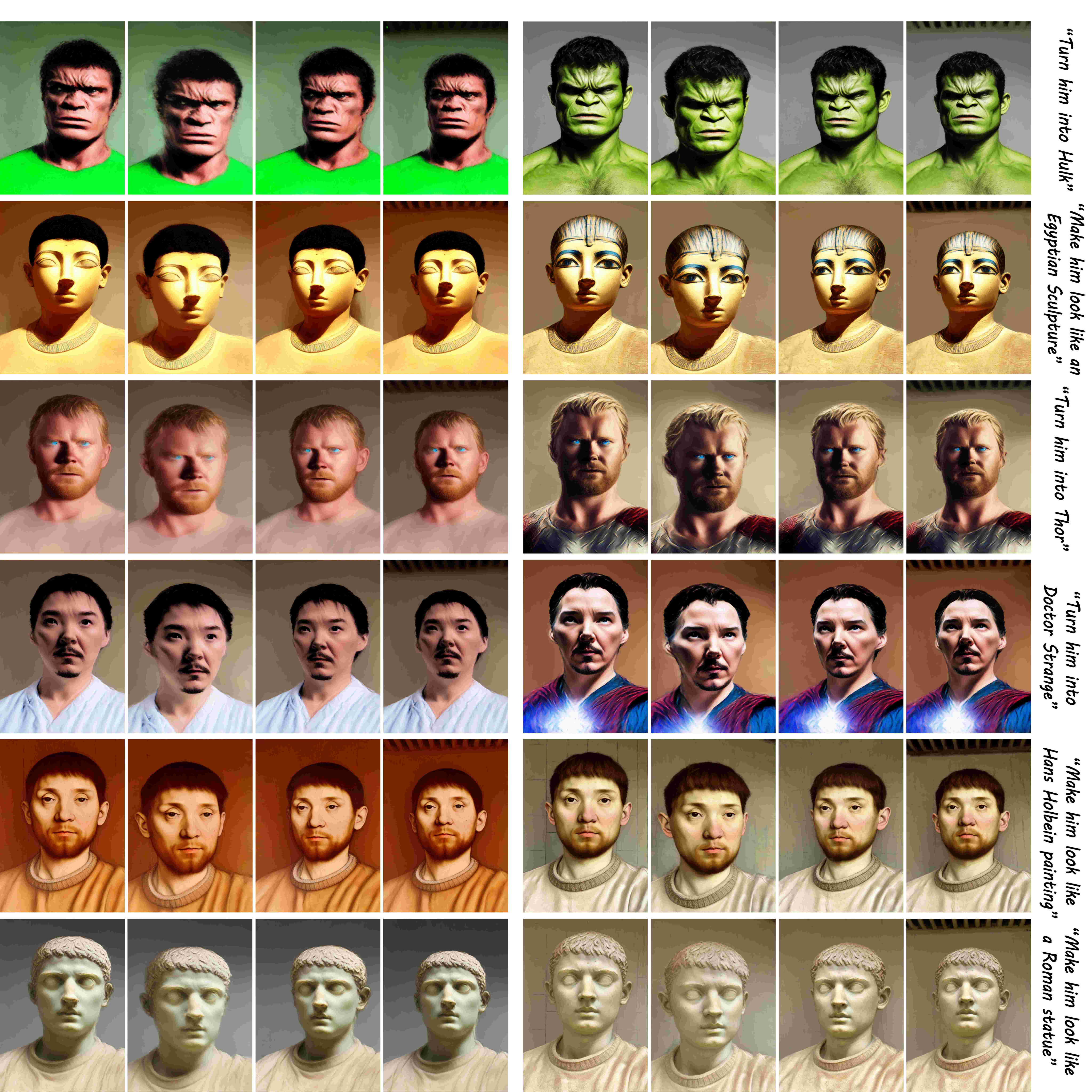}\\
	\caption{Qualitative comparison. Left: Instruct-Nerf2Nerf~\cite{haque2023instruct}. Right: Our SyncNoise.}
	\label{app:fangzhou} 
\end{figure}

We provide more qualitative comparisons with Instruct-Nerf2Nerf~\cite{haque2023instruct} in Fig.~\ref{app:fangzhou}. In comparison, our method has four advantages. (1) More details. The early stage of IN2N results in vastly different multi-view editing outputs, leading to blurry results. Our method, however, performs alignment on both noise and pixels, thereby retaining more details. (2) Higher controllability and less interference with the background. Our method introduces a soft mask to limit the editing region, resulting in fewer changes to the background. (3) Faster convergence. Our method can achieve better results with half the training time of IN2N. (4) Better geometric consistency. We align features on the U-Net, better ensuring the structural consistency across multiple views.

\subsection{Alignment of Different U-Net Layers}
\label{app:alignment}
\begin{figure}[!h]
	\centering 
	\includegraphics[scale=3.0]{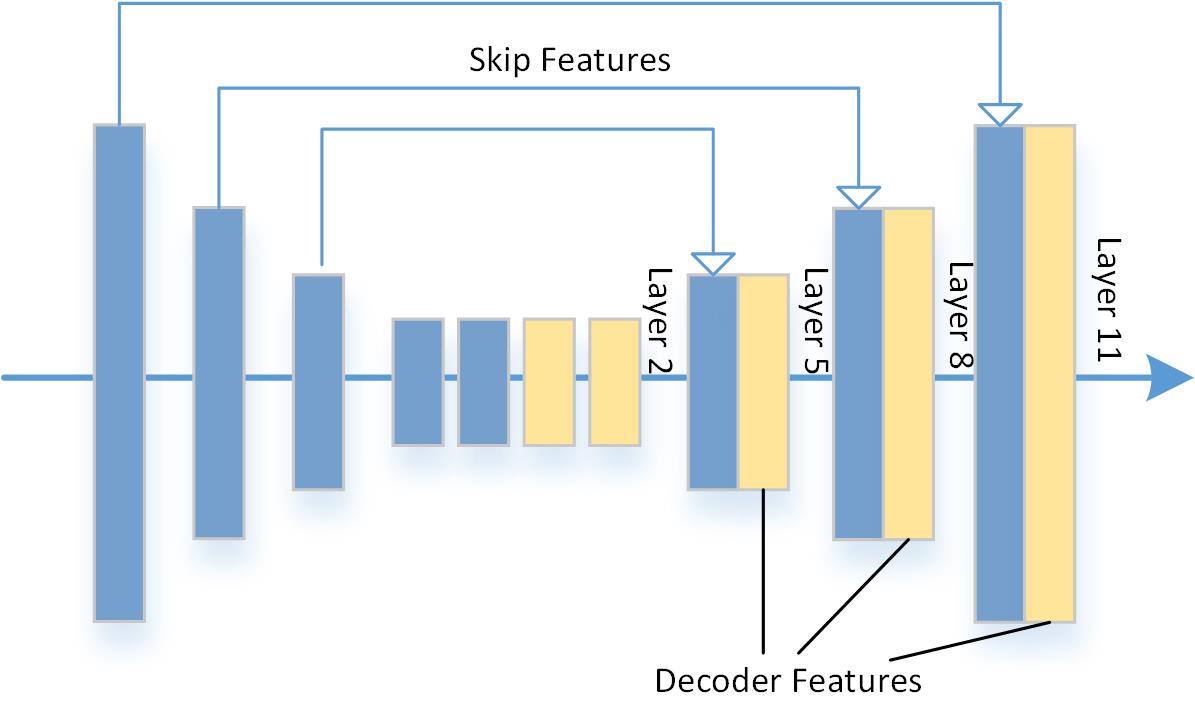}\\
	\caption{U-Net architecture.}
	\label{app:unet} 
\end{figure}

\begin{figure}[!h]
	\centering 
	\includegraphics[scale=0.18]{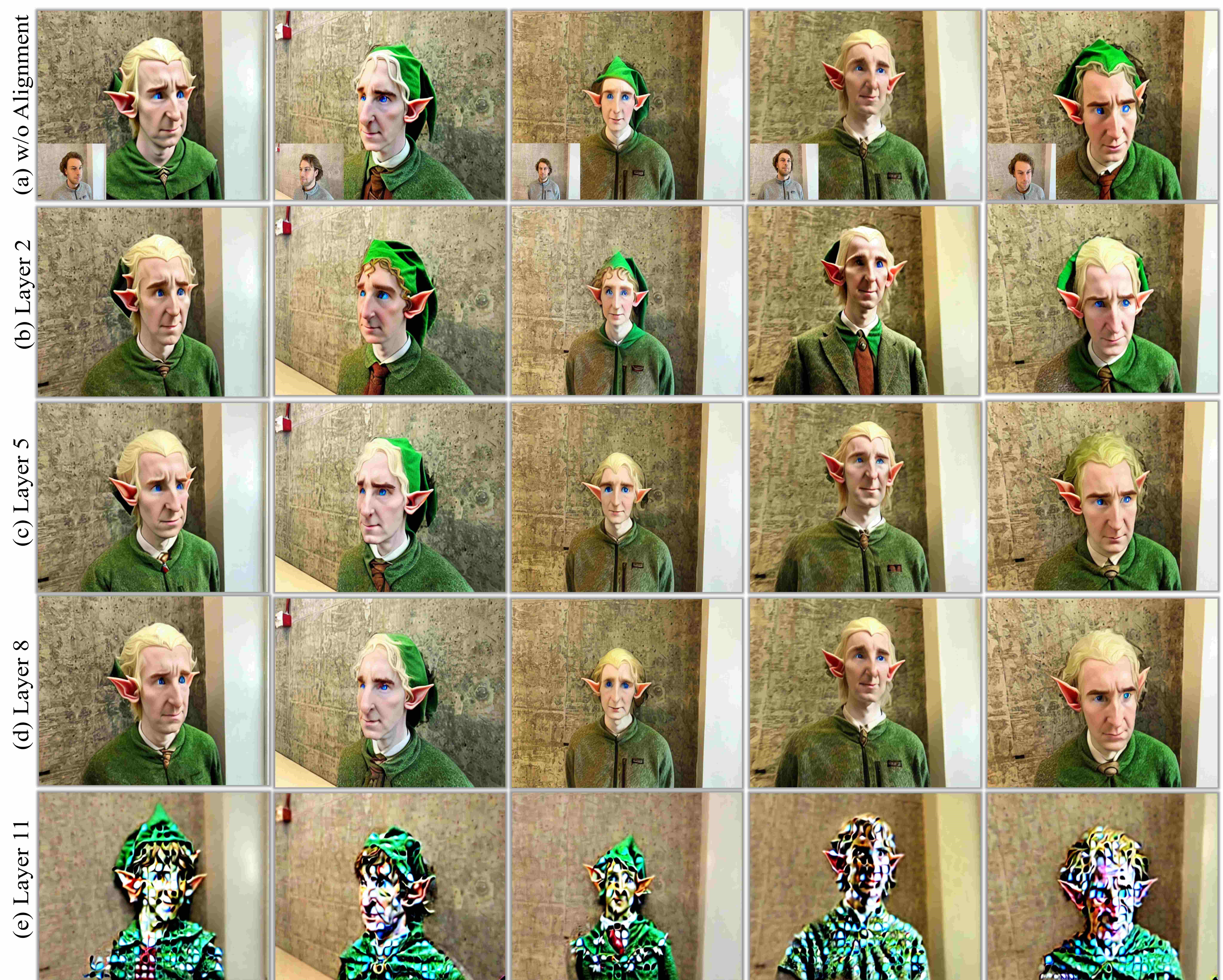}\\
	\caption{Multi-view editing results by enforcing the consistency on different layers of decoder features of U-Net.}
	\label{app:layer} 
\end{figure}

The U-Net architecture is shown in Fig.~\ref{app:unet}. We enforce the multi-view consistency on different layers of U-Net. As can be seen in Fig.~\ref{app:layer}, when performing alignment at Layer-5 and Layer-8, the consistency across multiple views is better preserved. Aligning at Layer-11, however, introduces many artifacts, as this disrupts the predicted noise distribution. Therefore, we ultimately chose to align at Layer-5 and Layer-8.

\end{document}